\documentclass{bmvc2k}

\title{Federated Learning for Face Recognition via Intra-subject Self-supervised Learning}

\addauthor{Hansol Kim}{hans.kim@kakaobank.com}{1,2}
\addauthor{Hoyeol Choi}{con.choi@kakaobank.com}{1}
\addauthor{Youngjun Kwak}{vivaan.yjkwak@kakaobank.com,yjk.kwak@kaist.ac.kr}{$\dagger$,1,3}

\addinstitution{
 KakaoBank Corp., South Korea
}
\addinstitution{
 Graduate School of Software Technology, Kookmin University, South Korea
}

\addinstitution{
 Department of Electrical Engineering, KAIST, South Korea
}

\runninghead{Hansol Kim, Hoyeol Choi, Youngjun Kwak}{KakaoBank Corp.}

\usepackage{color, colortbl}
\usepackage{xcolor}

\usepackage{multirow}
\usepackage{algorithm}
\usepackage{algpseudocode}
\usepackage{amssymb}
\usepackage{pdfrender}

\begin{document}

\maketitle

\begin{abstract}
Federated Learning (FL) for face recognition aggregates locally optimized models from individual clients to construct a generalized face recognition model. However, previous studies present two major challenges: insufficient incorporation of self-supervised learning and the necessity for clients to accommodate multiple subjects. To tackle these limitations, we propose FedFS (Federated Learning for personalized Face recognition via intra-subject Self-supervised learning framework), a novel federated learning architecture tailored to train personalized face recognition models without imposing subjects. Our proposed FedFS comprises two crucial components that leverage aggregated features of the local and global models to cooperate with representations of an off-the-shelf model. These components are (1) adaptive soft label construction, utilizing dot product operations to reformat labels within intra-instances, and (2) intra-subject self-supervised learning, employing cosine similarity operations to strengthen robust intra-subject representations. Additionally, we introduce a regularization loss to prevent overfitting and ensure the stability of the optimized model. To assess the effectiveness of FedFS, we conduct comprehensive experiments on the DigiFace-1M and VGGFace datasets, demonstrating superior performance compared to previous methods.
\end{abstract}

\section{Introduction}
\label{sec:intro}
Recent years have witnessed a burgeoning interest in safeguarding personal data, a concern further emphasized by Article 25 of the GDPR \cite{truong2021privacy}, which mandates heightened data protection measures throughout system development and prohibits the unauthorized collection of personal information. Consequently, safeguarding personal information during the training of deep-learning networks has emerged as a paramount concern.

Face recognition has garnered considerable attention due to its efficacy in identifying individuals. This method finds widespread application in user authentication and has even found integration into smartphones to safeguard personal information or financial transactions. However, a significant portion of face recognition models \cite{kim2022adaface, kim2020discface, li2021dynamic} are typically hosted on servers, necessitating the transmission of facial images from smartphones for authentication, which raises privacy concerns. To address this issue, the adoption of lightweight models directly on smartphones has been proposed. Nonetheless, limitations persist in training these models solely using public data. Consequently, there is a growing interest in training models to utilize users' facial data on their own devices while ensuring data privacy, with many studies leveraging federated learning methods garnering attention in this regard.

Federated learning is a method in which multiple clients join together to train a model with good performance while protecting personal information. FedFace \cite{aggarwal2021fedface} introduced a spread-out regularizer aimed at training a face recognition model within a federated learning framework. However, the process of dispersing the identity proxies received from clients in FedFace raises concerns regarding potential privacy violations. FedFR \cite{liu2022fedfr} prevented bias by training personalized models using public data, demonstrating promising performance among federated learning-based face recognition models. However, this approach necessitates clients to continuously receive public data, posing significant resource constraints, especially in on-device environments like mobile platforms where computational resources are severely limited. Additionally, FedFR proposed a new evaluation metric for personalizing performance, but this metric is far from real-world situations because the number of clients is too small and one client holds multiple identities.

To address these challenges, we propose FedFS, which trains generalized facial features and personalized face recognition model without leaking personal data outside each user's device in a federated learning environment. FedFS has three models: a pre-trained model trained with public data, a personalized model, and a global model, as well as two components: adaptive soft label construction utilizing dot product and intra-subject self-supervised learning employing cosine similarity, to reduce computational complexity and intra-class variation. Additionally, we introduce regularization loss to prevent bias in personalized models in heterogeneous data situations. We assume an actual authentication environment in which tens of thousands of clients participate in federated learning and evaluate personalizing performance using DigiFace-1M \cite{bae2023digiface} and VGGface \cite{cao2018vggface2} benchmark data. Our main contributions are summarized as follows:

\begin{figure}[]
\centerline{\includegraphics[width=\textwidth]{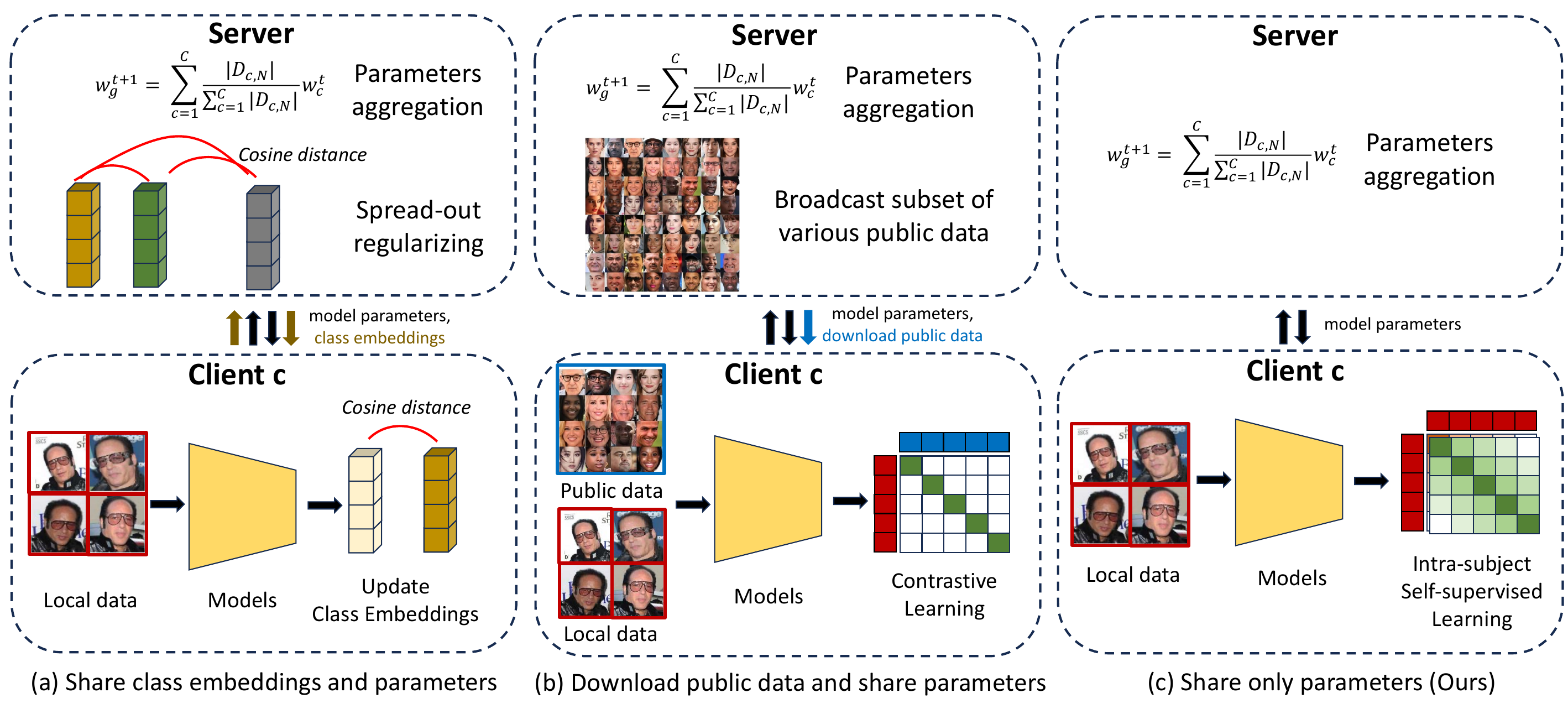}}
\caption{Pipelines of federated learning-based face recognition methods including our proposed method. (a) The server collects class embedding of client c (e.g. FedFace). (b) Client c continuously downloads public data from the server (e.g. FedFR). (c) Our proposed method(FedFS), client c performs intra-subject self-supervised learning without any additional work.}
\vspace{-0.3cm}
\label{fig:intro}
\end{figure}

\begin{itemize}
\item We propose FedFS, Federated Learning for personalized Face recognition via intra-subject Self-supervised learning framework. FedFS trains optimized facial features for each client and reduces intra-class variation by leveraging adaptive soft label construction utilizing dot product and intra-subject self-supervised learning employing cosine similarity while protecting users' data privacy.

\item Regularization loss is proposed to prevent bias in the performance of personalized models. Through this, FedFS solves the problem of easily falling into overfitting when training only with personal data, and trains indirectly generalized facial features.

\item Experiment results utilizing face recognition benchmarks like DigiFace-1M and VGGFace demonstrate that our proposed method outperforms previous approaches. Furthermore, we conducted training and evaluation with the assumption of 10,000 clients participating, each with only one identity, mirroring real-world conditions. This assumption is the first attempt in federated learning-based face recognition research.
\end{itemize}
\section{Related Works}
\textbf{Face Recognition.} Face recognition has seen a remarkable enhancement in performance through the utilization of large-scale data, identities, and models, sparking considerable interest \cite{wang2021deep, boutros2022sface, 10095535}. However, state-of-the-art models require a lot of resources, so the execution environment is often limited to servers with no resource limitations. In this case, personal information is violated because the actual authentication process requires facial data to be transmitted from the client to the server. In contrast to large-scale models, there is a growing body of research focusing on lightweight models \cite{alansari2023ghostfacenets, boutros2022pocketnet}. MobileFaceNet \cite{chen2018mobilefacenets} exemplifies one such lightweight face recognition model, boasting speeds that are more than twice as fast as MobileNetV2 \cite{sandler2018mobilenetv2}. However, enhancing the performance of pre-trained models like these, which utilize public data, proves challenging due to constraints imposed by model size and the inability to conduct additional training using user data. FedFS aims to address this limitation by enhancing recognition performance through personalized facial feature training while safeguarding personal information.
\\ \newline\textbf{Federated Learning.} Federated learning is attracting attention as a way to protect personal information. FedAvg \cite{mcmahan2017communication} is a method of creating a global model by calculating the average of the parameters of a local model trained using data from each client. Recently, research on personalized federated learning, which improves personalized performance by utilizing models customized to suit individual goals, is increasing. However, most research has only been conducted on small-scale datasets such as MNIST \cite{lecun1998gradient} and CIFAR-10 \cite{krizhevsky2009learning}. To solve these problems, research on face recognition has been conducted in federated learning environments such as FedFace \cite{aggarwal2021fedface} and FedFR \cite{liu2022fedfr}. FedFR simultaneously trained for generalizing performance and personalizing performance using public data. In contrast, we do not use public data directly, because utilizing the data requires the client's resources, which can be very taxing on the client's devices.
\\ \newline\textbf{Contrastive Learning.} Contrastive learning researchs achieve state-of-the-art results on learning image features \cite{chen2020simple, chen2020big, he2020momentum}. The main idea of contrastive learning is to diminish the distance between the features of the same identity of the images and increase the distance between the features of different identities of images. In the past, dot products were widely used in contrastive learning, but recently, cosine similarity has been widely used \cite{deng2019arcface, wang2018cosface}. This shift is attributed to the potential of dot products to yield large values based on the data, resulting in various issues such as inflated weight values \cite{thongtan2019sentiment}. However, geometrically speaking, cosine similarity solely concerns angular, rendering normalized data indistinguishable in magnitude. This means that cosine similarity is effective in maximizing inter-class variation, but shows poor performance in some cases \cite{steck2024cosine}. In contrast, the dot product is influenced not only by the angle but also by the magnitude, enabling differentiation even among data involving the same identifier. From this perspective, we aim to minimize intra-class variation by using dot product and cosine similarity simultaneously.

Contrastive learning is garnering significant attention due to its outstanding performance, and numerous studies applying federated learning and contrastive learning are underway. Unlike traditional contrastive learning approaches, in federated contrastive learning, clients can only have their data, so there are no other identities. To address this challenge, a variety of federated learning-based studies \cite{liu2022fedfr, kim2023protofl} are attracting much attention. In this paper, we focus on effectively learning individual features (positive data) without other identities (negative data) in a federated learning setup and propose regularization loss to prevent overfitting and bias.

\begin{algorithm}[!ht]
\caption{Procedure of FedFS}\label{alg:cap}
\begin{algorithmic}
\State Communication Round is $t$, $t \in \{0, ..., T\}$.
\State Initialize a server-global model parameters $w_g^0$.
\State Broadcast pre-trained model $\xi$ to all participating clients.

\State \textbf{Server executes}:
\For{$t = 0, ... , T$}
    \For{$c = 1, ... , C$}
        \State Send the server-global model parameters $w_g^t$ to client c
        \State $w_c^{t}$ and $|D_{c,N}|$ $\gets$ \textbf{ClientTraining($w_g^t$)}
    \EndFor
    \State $w_g^{t+1} = \sum_{c=1}^{C}\frac{|D_{c,N}|}{\sum_{c=1}^{C}|D_{c,N}|}w_c^{t}$
\EndFor
\\
\State \textbf{function} ClientTraining($w_g^t$):
\State $w_c^t$ $\gets$ $w_g^t$
\For{$i=1, ..., N$}
    \State $F_{total}(\psi, w_c, \theta_c)=\lambda*F_{insub}(\psi, w_c, \theta_c)+(1-\lambda)*F_{reg}(w_c, \theta_c)$
    \State $w_c^{t}$ $\gets$ $w_c^t - \eta\nabla_{w_c}F_{total}(\psi,w_c,\theta_c)$
    \State $\theta_c^{t}$ $\gets$ $\theta_c^t - \eta\nabla_{\theta_c}F_{total}(\psi,w_c,\theta_c)$
\EndFor
\State Calculate the number of the c client data $D_{c,N}$
\State \textbf{return} $w_c^t$ and $|D_{c,N}|$
\State \textbf{end function}
\end{algorithmic}
\end{algorithm}

\section{Proposed Method}
In this section, we propose a federated learning framework for personalized face recognition with intra-subject self-supervised learning and this flow is summarized in Algorithm \ref{alg:cap}. We will first describe the training environment and then explain in detail the training process proposed in this environment. Additionally, we demonstrate the convergence analysis for our proposed method in Appendix \ref{app:conv}.

\subsection{Problem Formulation}
We define the total number of participating clients in federated learning as $C$, and the specific client as $c$, $c\in\{1, ..., C\}$. Clients combine a personalized model, a pre-trained model, and a global model to collectively train their individual facial features. The personalized model has the same architecture as the global model. Each client has a training dataset $D_{C,N}=\{x_{c,i}, 1 \leq i \leq N, 1 \leq c \leq C\}$, where $N$ is the cardinality of the local data $D_{C,N}$. In a federated learning setup, the parameter server collects and aggregates the parameters of the global model from each client without sharing any private data. We adopt the commonly used FedAvg \cite{mcmahan2017communication} as our aggregate baseline method. This step is summarized as follows:
\begin{equation}
w_g^{t+1} = \sum_{c=1}^{C}\frac{|D_{c,N}|}{\sum_{c=1}^{C}|D_{c,N}|}w_c^{t}
\end{equation}
\newline where $t$ means $t$th communication round, $t \in \{0, ..., T\}$, $w_g$ is the parameters of server-global model, $w_c$ is the parameters of global model trained in personal device of client $c$ and $|D_{c,N}|$ is the number of samples on dataset $D_{c,N}$. After updating the parameters of the server-global model, the parameters are broadcast to all clients. Through this process, we can indirectly train generalized facial features.
\subsection{Intra-subject self-supervised learning}
\textbf{Intra-subject representations}. In intra-subject self-supervised learning, two major operations are performed simultaneously. 1) Training local information and reducing intra-class variation with intra-subject loss. 2) Preventing overfitting and bias with regularization loss. Considering the client's restriction to utilize only local data for privacy protection, each client trains the model using only positive data, excluding negative data. Under these conditions, the client $c$ performs operations with the global model, personalized model, and pre-trained model on input data $x_{c,i}$ and obtains the following results, respectively:
\begin{equation}
    r_{c,i} = \phi_c(x_{c,i}, w_{c}),\quad q_{c,i} = \phi_c(x_{c,i}, \theta_{c}),\quad v_{c,i} = \xi(x_{c,i}, \psi), \quad z_{c,i} = r_{c,i} \oplus q_{c,i}
\end{equation}
\newline where $w_c$ is the global model($\phi(w)$) parameters of client c, $\theta_c$ is the personalized model($\phi(\theta)$) parameters of client c and $\psi$ is the pre-trained model($\xi$) parameters. Subsequently, we obtain the intra-subject representation using the cosine similarity between the results and calculate the intra-subject loss value within the online-batch. We do not update $\psi$ parameters, and share $\theta_c$ parameters with the server. The process is as follows:

\begin{figure*}[]
\centerline{\includegraphics[width=1.0\textwidth]{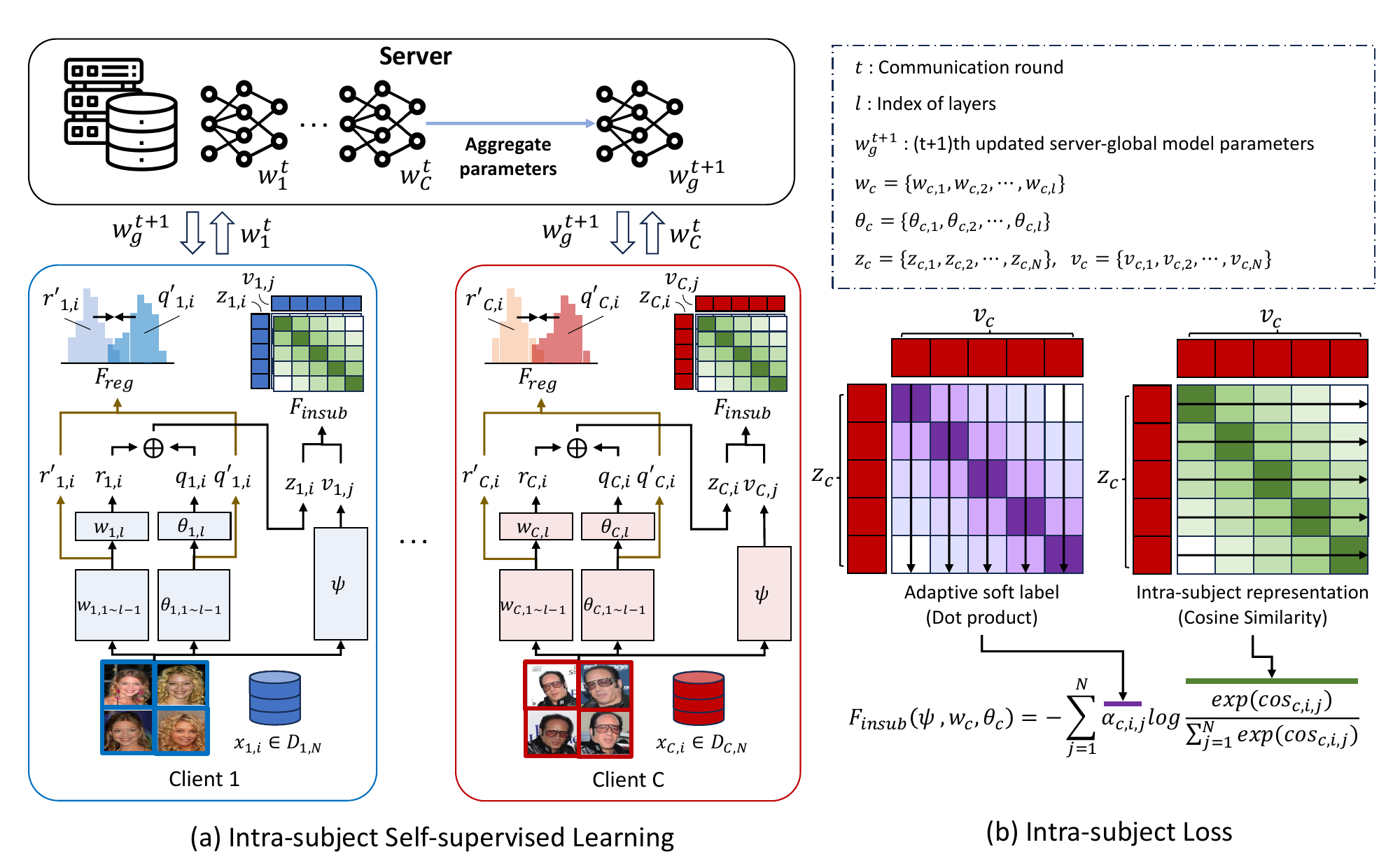}}
\caption{(a) is an overview of our proposed training process and (b) is the detailed process of intra-subject loss. The global model outputs two vectors and the personalized model also outputs two vectors. Using each output, we calculate regularization loss and create a $z_c$ vector. Intra-subject loss is measured using the $z_c$ vector and the output vector of the pre-trained model.}
\vspace{-0.5cm}
\label{fig:model_architecture}
\end{figure*}

\begin{equation}
    cos_{c,i,j} = 1 - \frac{z_{c,i}\cdot v_{c,j}}{||z_{c,i}||_2\cdot ||v_{c,j}||_2}
\end{equation}
\begin{equation}
F _{insub}(\psi, w_c, \theta_c) = - \sum_{j=1}^Ny_{c,j}log\frac{exp(cos_{c,i,j})}{\sum_{j=1}^Nexp(cos_{c,i,j})}\label{eq}
\end{equation}
\newline where $y_{c,j}$ means the class label, and $j$ has the same meaning as i $\{x_{c,j}, 1 \leq j \leq N \}$, but used to distinguish $v_c$ and $z_c$. The data $x_{c,i-1}$, $x_{c,i}$, and $x_{c,i+1}$ whthin the online-batch are all positive data, so each data have a high similarity to each other. However, due to the nature of cross entropy, $y_{c,j}$ is 0 except in cases where the input is the exactly same image within the online-batch. To address these limitations, our proposed method uses an adaptive soft label that reflects the correlation between all positive data to reformat labels within intra-instances and reduce intra-class variance, thereby more effectively training correlations for local data. \newline \newline \textbf{Adaptive soft label}. To obtain an adaptive soft label, we calculate the adaptive soft score $ass$ using the dot product. Additionally, we select the K ratio of the batch size in descending order to emphasize the correlation with the specific ratio. Afterward, instead of labels, we use the adaptive soft label. This process is as follows:

\begin{align}
    ass_{c,i,j} = z_{c,i} \cdot v_{c,j},\quad ASS_{c,i,j} \in \{ ass_{c,1,j}, ..., ass_{c,i,j} \}
\end{align}

\[
    \beta_{c,i,j}
\begin{cases}
    ass_{c,i,j} * \gamma, & \text{if } y_{c,j}=1\\
    ass_{c,i,j}, & \text{else if } \star \\
    0,              & \text{otherwise}
\end{cases}
\]\label{eq123}

\begin{equation}
    \alpha_{c,i,j} = (\frac{exp(\beta_{c,i,j})}{\sum_{i=1}^Nexp(\beta_{c,i,j})})^T
\end{equation}
\newline where $\star$ means $y_{c,j}=0$ and $ass_{c,i,j}$ is Top K in $ASS_{c,i,j}$. $ASS_{c,i,j}$ is adaptive soft score set of client $c$. The $\gamma$ is a positive number, and the K value is a value between 0 and 1. The $\gamma$ is a probability enhancement value for itself. As the value increases, the model focuses more on training the similarity to itself, and as it becomes smaller, the model trains by focusing on the similarity to surrounding vectors. In this paper, the K value is set to 4 and $\gamma$ is set to 2. Finally, the local facial features are trained by performing a cross-entropy operation using the adaptive soft label $\alpha$ instead of the previously used label value $y_{c,j}$. The process can be summarized as follows:

\begin{equation}
F_{insub}(\psi, w_c, \theta_c) = - \sum_{j=1}^N\alpha_{c,i,j}log\frac{exp(cos_{c,i,j})}{\sum_{j=1}^Nexp(cos_{c,i,j})}\label{eq:ssim}
\end{equation}
\newline \newline \textbf{Regularization loss}. Training only on local data without including negative data can easily lead to overfitting and biased results. To solve this problem, we perform regularizing between the global model that trains generalized facial features through sharing the parameters with the server and the personalized model, as follows:

\begin{align}
F_{reg}(w_c, \theta_c) = 1 - \frac{r'_{c,i}\cdot q'_{c,i}}{||r'_{c,i}||_2\cdot ||q'_{c,i}||_2}
\end{align}
\newline where $r'_{c,i}$ and $q'_{c,i}$ are the output vectors that do not pass through the last linear layer of the global model and personalized model, respectively. Finally, the intra-subject self-supervised learning process is summarized as follows.

\begin{equation}
F_{total}(\psi, w_c, \theta_c) = \lambda*F_{insub}(\psi, w_c, \theta_c) + (1-\lambda)*F_{reg}(w_c, \theta_c)
\label{eq:total}
\end{equation}
\newline where $\lambda$ is an objective weight value between 0 and 1. In this paper, $\lambda$ is set to 0.7.

\vspace{0.5cm}
\section{Experiments}
In this section, we demonstrate the performance of our proposed method through experiments. To evaluate the performance of each client's personalized face recognition model, we use an evaluation technique that arranges the evaluation data in a 1:N structure conducted in FedFR \cite{liu2022fedfr}. In addition, we check whether our proposed method reduces intra-class variation and ablation study in Appendix \ref{app:abl}.

\subsection{Experiment Setting}
We use MS-Celeb-1M \cite{guo2016ms} to train pre-trained models and share the data publicly in FedFR \cite{liu2022fedfr}. And we set MobileFaceNet \cite{chen2018mobilefacenets}, PocketNet \cite{boutros2022pocketnet}, GhostFaceNet \cite{alansari2023ghostfacenets} and MobileNetV2 \cite{sandler2018mobilenetv2} as pre-trained models. DigiFace-1M \cite{bae2023digiface} and VGGFace \cite{cao2018vggface2} are benchmark datasets for training and evaluation of personalized face recognition models. We use 80\% of the images in total for local client training and the remaining 20\% of the images for evaluation. Specifically, in each local client, the number of training data and evaluation data are 57/13 for DigiFace-1M and 100/13 for VGGFace, respectively. Also, DigiFace-1M and VGGFace have 10,000 and 8,673 identities, respectively, and each client has only one identity.

In this experiment, we employ 64-layer CNN architecture \cite{liu2017sphereface} as a global model and personalized model in the same way as FedFR. We add a linear layer to the last layer for intra-subject self-supervised learning. We use the SGD optimizer with a learning rate of 5e-3. Each client trains 2 local epochs and 5 communication rounds, and clients participating in training are selected randomly.
\begin{table}[!t]
\begin{center}
\resizebox{1.0\textwidth}{!}{\Huge
\begin{tabular}{|l|l|cc|cc|l|l|l|cc|cc|}
\cline{1-6} \cline{8-13}
\multirow{2}{*}{Pre-trained model} & \multirow{2}{*}{FL method} & \multicolumn{2}{c|}{DigiFace-1M}          & \multicolumn{2}{c|}{VGGFace}           &  & \multirow{2}{*}{Pre-trained model} & \multirow{2}{*}{FL method} & \multicolumn{2}{c|}{DigiFace-1M}          & \multicolumn{2}{c|}{VGGFace}          \\ \cline{3-6} \cline{10-13} 
                                   &                            & \multicolumn{1}{c|}{AUROC}  & \%      & \multicolumn{1}{c|}{AUROC}  & \%       &  &                                    &                            & \multicolumn{1}{c|}{AUROC}  & \%      & \multicolumn{1}{c|}{AUROC}  & \%      \\ \cline{1-6} \cline{8-13} 
MobileFaceNet                      & -                          & \multicolumn{1}{c|}{0.8248} & -       & \multicolumn{1}{c|}{0.8921} & -        &  & PocketNet                          & -                          & \multicolumn{1}{c|}{0.9128} & -       & \multicolumn{1}{c|}{0.9806} & -       \\ \cline{3-6} \cline{10-13} 
                                   & FedFace                    & \multicolumn{1}{c|}{0.5001} & -60.6\% & \multicolumn{1}{c|}{0.5488} & -61.51\% &  &                                    & FedFace                    & \multicolumn{1}{c|}{0.4998} & -54.7\% & \multicolumn{1}{c|}{0.5865} & -59.8\% \\ \cline{3-6} \cline{10-13} 
                                   & FedFR                      & \multicolumn{1}{c|}{0.8270} & +0.2\%  & \multicolumn{1}{c|}{0.9477} & +6.2\%   &  &                                    & FedFR                      & \multicolumn{1}{c|}{0.9637} & +5.5\%  & \multicolumn{1}{c|}{0.9875} & +0.7\%  \\ \cline{3-6} \cline{10-13} 
                                   & \cellcolor[HTML]{C0C0C0}\textbf{FedFS(Ours)}                & \multicolumn{1}{c|}{\cellcolor[HTML]{C0C0C0}\textbf{0.9629}} & \cellcolor[HTML]{C0C0C0}\textbf{+16.7\%} & \multicolumn{1}{c|}{\cellcolor[HTML]{C0C0C0}\textbf{0.9794}} & \cellcolor[HTML]{C0C0C0}\textbf{+9.7\%}   &  &                                    & \cellcolor[HTML]{C0C0C0}\textbf{FedFS(Ours)}                & \multicolumn{1}{c|}{\cellcolor[HTML]{C0C0C0}\textbf{0.9794}} & \cellcolor[HTML]{C0C0C0}\textbf{+7.2\%}  & \multicolumn{1}{c|}{\cellcolor[HTML]{C0C0C0}\textbf{0.9934}} & \cellcolor[HTML]{C0C0C0}\textbf{+1.3\%}  \\ \cline{1-6} \cline{8-13} 
GhostFaceNets                      & -                          & \multicolumn{1}{c|}{0.9612} & -       & \multicolumn{1}{c|}{0.9885} & -        &  & MobileNetV2                        & -                          & \multicolumn{1}{c|}{0.9339} & -       & \multicolumn{1}{c|}{0.9645} & -       \\ \cline{3-6} \cline{10-13} 
                                   & FedFace                    & \multicolumn{1}{c|}{0.5106} & -53.1\% & \multicolumn{1}{c|}{0.5905} & -59.7\%  &  &                                    & FedFace                    & \multicolumn{1}{c|}{0.5055} & -54.1\% & \multicolumn{1}{c|}{0.5542} & -57.4\% \\ \cline{3-6} \cline{10-13} 
                                   & FedFR                      & \multicolumn{1}{c|}{0.9644} & +0.3\%  & \multicolumn{1}{c|}{0.9929} & +0.4\%   &  &                                    & FedFR                      & \multicolumn{1}{c|}{0.9588} & +2.6\%  & \multicolumn{1}{c|}{0.9876} & +2.3\%  \\ \cline{3-6} \cline{10-13} 
                                   & \cellcolor[HTML]{C0C0C0}\textbf{FedFS(Ours)}                & \multicolumn{1}{c|}{\cellcolor[HTML]{C0C0C0}\textbf{0.9944}} & \cellcolor[HTML]{C0C0C0}\textbf{+3.4\%}  & \multicolumn{1}{c|}{\cellcolor[HTML]{C0C0C0}\textbf{0.9943}} & \cellcolor[HTML]{C0C0C0}\textbf{+0.5\%}   &  &                                    & \cellcolor[HTML]{C0C0C0}\textbf{FedFS(Ours)}                & \multicolumn{1}{c|}{\cellcolor[HTML]{C0C0C0}\textbf{0.9647}} & \cellcolor[HTML]{C0C0C0}\textbf{+3.3\%}  & \multicolumn{1}{c|}{\cellcolor[HTML]{C0C0C0}\textbf{0.9922}} & \cellcolor[HTML]{C0C0C0}\textbf{+2.8\%}  \\ \cline{1-6} \cline{8-13} 
\end{tabular}
}
\end{center}
\caption{AUROC of various federated learning methods on DigiFace-1M and VGGFace. Each method uses MobileFaceNet, PocketNet, GhostFaceNet, and MobileNetV2 as a pre-trained model to measure AUROC and the AUROC increase/decrease rate compared to the pre-trained model.}
\label{tab:tab_auroc}
\end{table}

\begin{figure*}[!t]
\centerline{\includegraphics[width=1.0\textwidth]{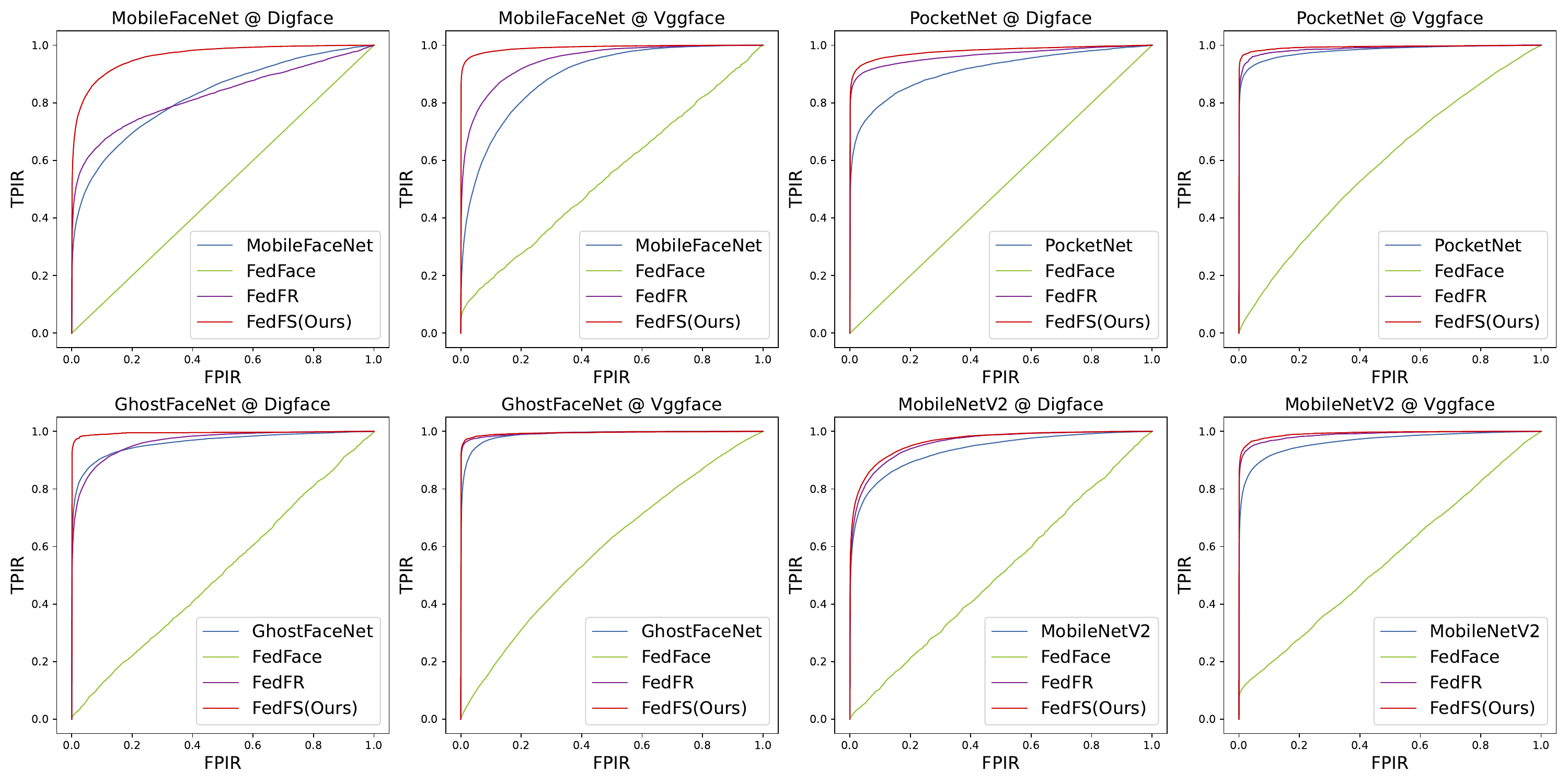}}
\caption{Each graph represents the ROC curve against the pre-trained model @ benchmark and the ROC curve against federated learning methods using the pre-trained model.}
\label{fig:model_roc}
\end{figure*}

\subsection{Experiment Results}
We conduct experiments to analyze how much performance is improved compared to the pre-trained model using FedFace \cite{aggarwal2021fedface}, FedFR \cite{liu2022fedfr}, and our proposed method, FedFS. We use four pre-trained models: MobileFaceNet \cite{chen2018mobilefacenets}, PocketNet \cite{boutros2022pocketnet}, GhostFaceNet \cite{alansari2023ghostfacenets}, and MobileNetV2 \cite{sandler2018mobilenetv2}. The participation rate of all federated learning algorithms is 0.7, and we calculate AUROC and the percentage improvement based on the pre-trained model. These results are summarized in Table \ref{tab:tab_auroc}, and the ROC Curve graph under the same conditions is shown in Figure \ref{fig:model_roc}.

As a result of Table \ref{tab:tab_auroc} and Fugure \ref{fig:model_roc}, we can see that most models show good performance compared to the pre-trained model, but FedFS has the best performance. Because pre-trained models are trained based on large amounts of public data, they are difficult to retrain, and collecting the user data causes privacy issues. Additionally, in the case of the FedFace and FedFR, the performance improvement is not high because they assume a small number of clients and a 1.0 participation rate rather than a large number of clients. On the other hand, our proposed FedFS effectively trains facial features and reduces intra-class variation through intra-subject self-supervised learning using only local data without violating personal information, and shows that significantly improves personalized face recognition performance.
\begin{table}[]
\begin{center}
\resizebox{\textwidth}{!}{\Huge
\begin{tabular}{c|l|ccc|ccc}
\hline
\multirow{2}{*}{Participation rates} & \multirow{2}{*}{Method} &               & DigiFace-1M        &                 &               & VGGFace        &                 \\ \cline{3-8} 
                                  &                         & TPIR@FPIR=0.1 & TPIR@FPIR=0.01 & TPIR@FPIR=0.001 & TPIR@FPIR=0.1 & TPIR@FPIR=0.01 & TPIR@FPIR=0.001 \\ \hline
\multirow{3}{*}{0.01}             & FedFace                 & 0             & 0              & 0               & 0             & 0              & 0               \\
                                  & FedFR                   & 0.4139        & 0.2133         & 0.0866          & 0.7461        & 0.5464         & 0.3347          \\
                                  & \cellcolor[HTML]{C0C0C0}\textbf{FedFS(Ours)}             & \cellcolor[HTML]{C0C0C0}\textbf{0.6623}        & \cellcolor[HTML]{C0C0C0}\textbf{0.3142}         & \cellcolor[HTML]{C0C0C0}\textbf{0.1397}          & \cellcolor[HTML]{C0C0C0}\textbf{0.9680}        & \cellcolor[HTML]{C0C0C0}\textbf{0.9092}         & \cellcolor[HTML]{C0C0C0}\textbf{0.829}           \\ \hline
\multirow{3}{*}{0.1}              & FedFace                 & 0             & 0              & 0               & 0             & 0              & 0               \\
                                  & FedFR                   & 0.5239        & 0.2333         & 0.1113          & 0.8154        & 0.5917         & 0.3513          \\
                                  & \cellcolor[HTML]{C0C0C0}\textbf{FedFS(Ours)}             & \cellcolor[HTML]{C0C0C0}\textbf{0.8383}        & \cellcolor[HTML]{C0C0C0}\textbf{0.6057}         & \cellcolor[HTML]{C0C0C0}\textbf{0.3966}          & \cellcolor[HTML]{C0C0C0}\textbf{0.9688}        & \cellcolor[HTML]{C0C0C0}\textbf{0.9149}         & \cellcolor[HTML]{C0C0C0}\textbf{0.8391}          \\ \hline
\multirow{3}{*}{0.3}              & FedFace                 & 0             & 0              & 0               & 0             & 0              & 0               \\
                                  & FedFR                   & 0.6623        & 0.3066         & 0.18            & 0.8308        & 0.6191         & 0.3759          \\
                                  & \cellcolor[HTML]{C0C0C0}\textbf{FedFS(Ours)}             & \cellcolor[HTML]{C0C0C0}\textbf{0.8603}        & \cellcolor[HTML]{C0C0C0}\textbf{0.6164}         & \cellcolor[HTML]{C0C0C0}\textbf{0.4113}          & \cellcolor[HTML]{C0C0C0}\textbf{0.9765}        & \cellcolor[HTML]{C0C0C0}\textbf{0.925}          & \cellcolor[HTML]{C0C0C0}\textbf{0.8541}          \\ \hline
\multirow{3}{*}{0.5}              & FedFace                 & 0             & 0              & 0               & 0.0412        & 0              & 0               \\
                                  & FedFR                   & 0.7384        & 0.3666         & 0.2218          & 0.9152        & 0.7474         & 0.4743          \\
                                  & \cellcolor[HTML]{C0C0C0}\textbf{FedFS(Ours)}             & \cellcolor[HTML]{C0C0C0}\textbf{0.8752}        & \cellcolor[HTML]{C0C0C0}\textbf{0.6567}         & \cellcolor[HTML]{C0C0C0}\textbf{0.4586}          & \cellcolor[HTML]{C0C0C0}\textbf{0.9766}        & \cellcolor[HTML]{C0C0C0}\textbf{0.9266}         & \cellcolor[HTML]{C0C0C0}\textbf{0.8586}          \\ \hline
\multirow{3}{*}{0.7}              & FedFace                 & 0.0912        & 0.0103         & 0               & 0.1231        & 0.0195         & 0.0081          \\
                                  & FedFR                   & 0.7451        & 0.4743         & 0.3141          & 0.9315        & 0.8296         & 0.6026          \\
                                  & \cellcolor[HTML]{C0C0C0}\textbf{FedFS(Ours)}             & \cellcolor[HTML]{C0C0C0}\textbf{0.8905}        & \cellcolor[HTML]{C0C0C0}\textbf{0.6927}         & \cellcolor[HTML]{C0C0C0}\textbf{0.5072}          & \cellcolor[HTML]{C0C0C0}\textbf{0.9786}        & \cellcolor[HTML]{C0C0C0}\textbf{0.9337}         & \cellcolor[HTML]{C0C0C0}\textbf{0.8721}          \\ \hline
\end{tabular}
}
\end{center}
\caption{Performance comparison federated learning methods on DigiFace-1M and VGGFace benchmarks. Our proposed method shows the best performance in various participation rates environments.}
\label{tab:tab_compare}
\end{table}

\subsection{Performance with various participation rates}
Additionally, we compare the performance of the federated learning method using true positive identification rates (TPIR) at different false positive identification rates (FPIR) for 1:N identification protocol \cite{liu2022fedfr}. Specifically, we calculate the average TPIR of all clients based on FPIR 0.1, 0.01, and 0.001. The pre-trained model is MobileFaceNet \cite{chen2018mobilefacenets}, and we set various participation rates: 0.01, 0.1, 0.3, 0.5, and 0.7. According to the experimental results Table \ref{tab:tab_compare}, the performance of the proposed FedFS shows the best performance in all fields. Through this, FedFS, training using the intra-subject self-supervised learning method, is less affected by the participation rates compared to the previously federated learning methods FedFace \cite{aggarwal2021fedface} and FedFR \cite{liu2022fedfr}. \vspace{-0.2cm}
\section{Conclusion}
We proposed FedFS, a federated learning framework to train optimized facial features for each client by using intra-subject self-supervised learning while protecting personal information. Through intra-subject self-supervised learning, we could effectively learn a user's facial features and reduce intra-class variation by simultaneously leveraging dot product and cosine similarities among personal data, resulting in improved recognition performance compared to previous federated learning methods. We believe that FedFS could be applied to various federated face recognition tasks.
\bibliography{egbib}

\clearpage
\appendix
\addcontentsline{toc}{section}{Appendices}
\section*{Appendix}
\setcounter{equation}{0}
\renewcommand{\theequation}{\Alph{section}.\arabic{equation}}
\section{Congervence Analysis}\label{app:conv}
In this section, we analyze the convergence of FedFS. To do the analysis, we first need to make some preliminary definitions. The point where the local model is trained is designated as e. (For example, $w^{e+1}$ means the model parameter that has completed the $e+1$th training.) We denote the loss function of FedFS as $F$. We do not display $\psi$ separately, because the parameter is not updated.
\newline\newline\textbf{Assumption 1. Lipschitz Smoothness.}\\
If the gradient of the local model of any client c is L-Lipschitz smooth, the following formula holds.

\begin{equation}
||\nabla_w F(w, \theta^1) - \nabla_w F(w, \theta^2) ||_2 <= L||\theta^{1}-\theta^{2}||_2
\end{equation}
\begin{equation}
||\nabla_\theta F(w^1,\theta) - \nabla_\theta F(w^2,\theta) ||_2 <= L||w^{1}-w^{2}||_2
\end{equation}
\newline\textbf{Assumption 2. Unbiased Gradient and Bounded Variance}
\newline The $w$ and $\theta$ parameters each use SGD as an optimization function, so they each have unbiased and bounded variance. The parameter update process using SGD is as follows:

\begin{equation}
w^{e+1} = w^{e} - \eta\nabla_wF(w^e, \theta^e)
\end{equation}
\begin{equation}
\theta^{e+1} = \theta^{e} - \eta\nabla_\theta F(w^e, \theta^e)
\end{equation}
\newline Assuming that the amount of change in the parameter is within a certain range, the conditions are as follows:

\begin{equation}
||w^{e} - w^*||_2 <= \epsilon_w
\end{equation}
\begin{equation}
||\theta^{e} - \theta^*||_2 <= \epsilon_\theta
\end{equation}
\newline where $\epsilon_w$ and $\epsilon_\theta$ are positive value. With this assumption, updates to each parameter occur randomly within a certain range.
\newline\newline \textbf{Theorem 1. Convergence analysis}\\
Based on Assumption 1 and Assumption 2, a convergence analysis of FedFS can be performed as follows:

\begin{align}
||w^{e+1} - w^*||_2 <= \epsilon_w - \eta L\epsilon_\theta \\
||\theta^{e+1} - \theta^*||_2 <= \epsilon_\theta - \eta L\epsilon_w
\end{align}
\newline Based on the above analysis, we confirm that the parameters of FedFS converge within the real number range.
\newline\newline \textbf{Proof 1.}\\
First, the change in the objective function is estimated using the distance between parameters as follows.
\begin{align}
    ||w^{e+1} - w^*||_2 = ||w^e - \eta\nabla_w F(w^e, \theta^e) - w^*||_2 \\
    = ||w^e - w^* - \eta\nabla_w F(w^e, \theta^e)||_2 \\
    = ||w^e - w^* - \eta(\nabla_w F(w^e, \theta^e) - \nabla_w F(w^*, \theta^*))||_2 \\
    <= ||w^e - w^*||_2 - \eta||(\nabla_w F(w^e, \theta^e) - \nabla_w F(w^*, \theta^*)||_2 \\ 
    <= ||w^e - w^*||_2 - \eta L||\theta^e-\theta^*||_2
\end{align}
\newline Similar to the expansion process above, the same process is repeated for $w$ to obtain the follows:

\begin{equation}
||\theta^{e+1} - \theta^*||_2 <= ||\theta^e - \theta^*||_2 - \eta L||w^e-w^*||_2    
\end{equation}
\newline Now, based on Assumption 2, we develop the following:

\begin{align}
||w^{e+1} - w^*||_2 <= ||w^e - w^*||_2 - \eta L||\theta^e-\theta^*||_2 \\
<= \epsilon_w - \eta L\epsilon_\theta \\
||\theta^{e+1} - \theta^*||_2 <= ||\theta^e - \theta^*||_2 - \eta L||w^e-w^*||_2 \\
<= \epsilon_\theta - \eta L\epsilon_w
\end{align}
\newline Through this, Theorem 1 was proven.
\setcounter{figure}{0}
\setcounter{table}{0}
\renewcommand{\thetable}{\Alph{section}.\arabic{table}}
\renewcommand{\thefigure}{\Alph{section}.\arabic{figure}}
\section{Ablation Studies}\label{app:abl}

\begin{table}[!ht]
\begin{center}
\resizebox{1.0\textwidth}{!}{\Huge
\begin{tabular}{lcc|cc}
\hline
\multicolumn{1}{c|}{\multirow{2}{*}{\textbf{Setup}}} & \multicolumn{2}{c|}{\textbf{Modules}}                                                  & \textbf{DigiFace-1M} & \textbf{VGGFace} \\ \cline{2-5} 
\multicolumn{1}{c|}{}                                & \multicolumn{1}{l|}{Regularize loss}             & \multicolumn{1}{l|}{Adaptive soft label}   & TPIR@FPIR=0.001  & TPIR@FPIR=0.001  \\ \hline
\multicolumn{3}{l|}{Centrally trained with PocketNet}                                                                                         & 0.9128             & 0.9806             \\ \hline
\multicolumn{1}{l|}{A}                               & \texttimes & \texttimes       & 0.9051             & 0.9645             \\
\multicolumn{1}{l|}{B}                               & \checkmark & \texttimes     & 0.9537             & 0.9902             \\
\multicolumn{1}{l|}{Ours(C)}                         & \checkmark & \checkmark     & 0.9794             & 0.9934             \\ \hline
\end{tabular}
}
\end{center}
\caption{The TPIR performance for ablation studies}
\label{tab:tab_abla}
\end{table}


\begin{figure*}[!ht]
\centerline{\includegraphics[width=1.0\textwidth]{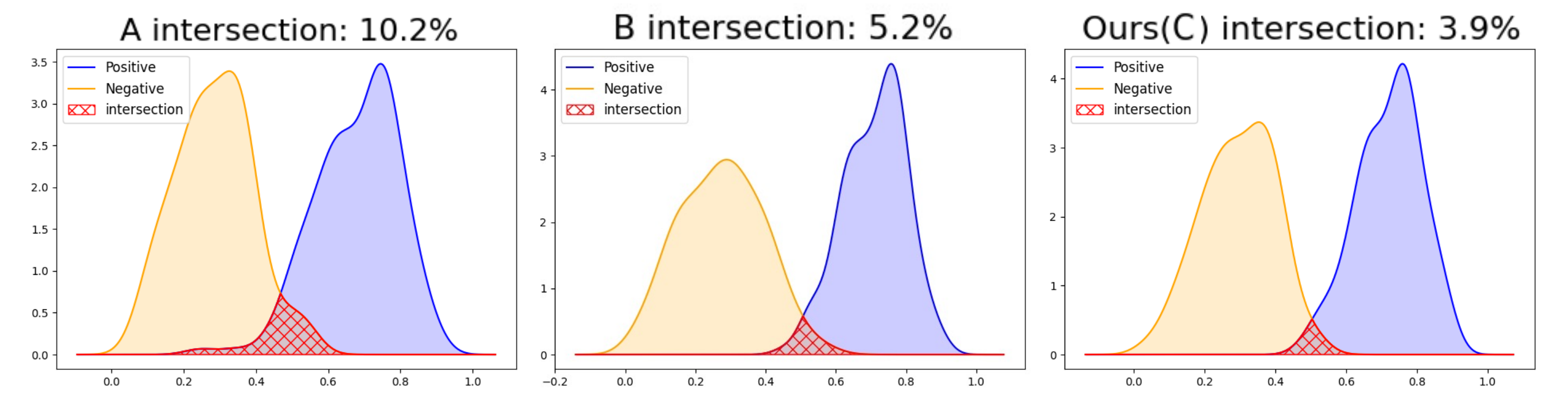}}
\caption{Average similarity distribution of clients participated in federated learning}
\label{fig:fig_abla}
\end{figure*}

Intra-subject self-supervised learning learning defined in the Equation 9, improves performance of personalized face recognition compared to previous approaches. We go further and check whether the proposed method reduces intra-class variation and analyze how the method affects performance. We set the participation rate at 0.7, and use PocketNet [5] as the pre-trained model. As shown in Figure \ref{fig:fig_abla} and Table \ref{tab:tab_abla}, we can see that the intra-subject self-supervised learning method considering correlation shows superior performance compared to using the general entropy learning method and regularize loss has a significant impact on performance by preventing overfitting and bias. In particular, as shown in the figure \ref{fig:fig_abla}, the proposed method has the smallest intersection of the positive similarity area and the negative similarity area. Through these results, we confirm that intra-subject self-supervised learning reduces intra-class variance.


\end{document}